\documentclass[letterpaper, 10 pt, conference]{ieeeconf}  

\IEEEoverridecommandlockouts                              

\usepackage{graphicx} 

\usepackage{amsmath} 
\usepackage{amssymb} 

\usepackage{algorithm}
\usepackage{algorithmic}
\usepackage{subfig} 

\usepackage{multirow}
\usepackage{booktabs}
\usepackage{cite}

\usepackage{xcolor} 

\title{\LARGE \bf
Q-Guided Stein Variational Model Predictive Control \\via RL-informed Policy Prior
}

\author{Shizhe Cai$^{1}$, Zeya Yin$^{1}$, Jayadeep Jacob$^{1}$ and Fabio Ramos$^{1,2}$
\thanks{$^{1}$School of Computer Science, The University of Sydney, Australia. Email: \{shizhe.cai, zyin0757, jayadeep.jacob, fabio.ramos\}@sydney.edu.au}
\thanks{$^{2}$NVIDIA, USA}
\thanks{Corresponding author: shizhe.cai@sydney.edu.au}
}

\begin{document}

\maketitle
\thispagestyle{empty}
\pagestyle{empty}

\begin{abstract}
Model Predictive Control (MPC) enables reliable trajectory optimization under dynamics constraints, but often depends on accurate dynamics models and carefully hand-designed cost functions. Recent learning-based MPC methods aim to reduce these modeling and cost-design burdens by learning dynamics, priors, or value-related guidance signals. Yet many existing approaches still rely on deterministic gradient-based solvers (e.g., differentiable MPC) or parametric sampling-based updates (e.g., CEM/MPPI), which can lead to mode collapse and convergence to a single dominant solution. We propose \textbf{Q-SVMPC}, a Q-guided Stein variational MPC method with an RL-informed policy prior, which casts learning-based MPC as trajectory-level posterior inference and refines trajectory particles via SVGD under learned soft Q-value guidance to explicitly preserve diverse solutions. Experiments on navigation, robotic manipulation, and a real-world fruit-picking task show competitive learning efficiency, strong final performance, and training stability compared with MPC, model-free RL, and learning-based MPC baselines.
\end{abstract}

\section{Introduction}
Model Predictive Control (MPC) provides reliable trajectory optimization under system constraints via receding-horizon re-planning. However, classical MPC often depends on accurate dynamics models and carefully designed cost functions, which can be difficult to obtain for complex robotic tasks. To mitigate these issues, recent work \cite{PETS, PaETS, ACMPC, AC4MPC, TDMPC, POLO, RL-Driven-MPPI, wang2025residualmppi} augments MPC with data-driven learning, including reinforcement learning (RL), to reduce modeling and cost-design burdens and improve planning quality. Other approaches enhance MPC in different ways: some learn cost or value-related functions to provide informative optimization signals~\cite{ACMPC, AC4MPC, TDMPC, POLO}, others learn system dynamics to improve prediction under uncertainty~\cite{PETS, PaETS}, and some learn better initialization or priors to warm-start trajectory optimization~\cite{RL-Driven-MPPI, wang2025residualmppi}. By injecting learned components into the MPC pipeline, these methods can improve planning quality and robustness to model errors in challenging robotic tasks.

Despite their promise, existing approaches exhibit limitations in how trajectories are represented and optimized. Some rely on deterministic gradient-based solvers \cite{ACMPC} (e.g., differentiable MPC \cite{diff_mpc}), which optimize a single trajectory at each step. Others adopt sampling-based planners \cite{PETS, PaETS, TDMPC, RL-Driven-MPPI, wang2025residualmppi} such as the Cross-Entropy Method (CEM) and Model Predictive Path Integral (MPPI), which iteratively refit a parametric distribution, typically from a Gaussian distribution. In both cases, trajectory optimization approximates a single high-return solution, which may collapse to a dominant trajectory and fail to preserve multiple feasible trajectories.

\begin{figure}[!t]  
    \centering
    \includegraphics[width=0.49\textwidth]{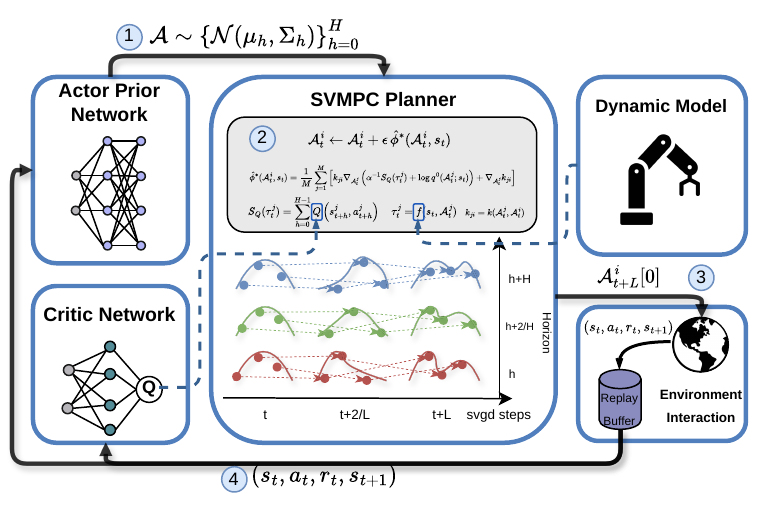}%
    \caption{\textbf{Overview of Q-SVMPC.} Q-SVMPC combines an RL-informed policy prior, model-predictive rollouts, and SVGD-based non-parametric trajectory refinement guided by learned soft Q-values.}
    \label{fig:model}
\end{figure}

Recent works formulate control, particularly MPC, as a probabilistic inference problem~\cite{RLCtrInf,honda2025mpcprobabilistic}. Under this perspective, finding an optimal policy becomes equivalent to performing approximate inference over a trajectory posterior defined by system dynamics and optimality variables. Methods such as Stein variational MPC (SVMPC) \cite{SVMPC} and DuSt-MPC \cite{DustMPC} adopt this formulation and employ SVGD for posterior approximation. However, as discussed in \cite{honda2025mpcprobabilistic}, key challenges remain in specifying appropriate prior distributions and in the manual design of task-dependent cost functions.

To address these limitations, we propose \textbf{Q-SVMPC} (Fig.~\ref{fig:model}), short for \textit{Q-Guided Stein Variational Model Predictive Control}. Q-SVMPC embeds non-parametric trajectory posterior refinement into a learning-guided MPC framework. Instead of relying on manually engineered costs, Q-SVMPC uses learned soft Q-values to define an optimality likelihood for trajectory inference. At each timestep, Q-SVMPC performs Bayesian inference conditioned on an RL-informed policy prior and refines trajectory particles using SVGD, explicitly preserving particle diversity while steering trajectories toward high-value regions. The first action of the refined sequence is executed, and the resulting trajectories provide improved learning signals for updating both the policy prior and the soft Q-function (following~\cite{SAC}), leading to more stable and sample-efficient learning.

We evaluate Q-SVMPC on 2D particle navigation, several robotic manipulation tasks, and a real-world fruit-picking scenario. The results show strong final performance across the evaluated benchmarks, with the largest gains observed on obstacle-rich reaching and contact-intensive manipulation tasks. Ablation studies further validate the effectiveness of Q-guided optimization and analyze the influence of horizon length, prior type, and the use of analytical versus learned dynamics. In summary, the Q-SVMPC framework brings the following contributions to the field:
\begin{itemize}
\item{A formulation of learning-guided MPC as trajectory-level posterior inference, using an RL-informed policy prior and learned soft Q-values as an optimality likelihood, and performing non-parametric posterior refinement via SVGD.}
\item{A theoretical connection between Soft Actor-Critic (SAC) and SVGD through the soft Q-value, together with an extension to trajectory-level inference, enabling SVGD-based optimization in learning-guided MPC.}
\item{Demonstration of Q-SVMPC on navigation, robotic manipulation, and a real-world fruit-picking task, showing competitive learning efficiency and improved final performance on the more challenging obstacle-rich and contact-intensive tasks.}
\end{itemize}

\section{Related Work}
\subsection{Learning-based MPC and MPC-in-the-loop Methods}
Model Predictive Control (MPC) has been increasingly combined with learned components to improve planning quality and data efficiency, giving rise to a family of learning-based MPC methods that integrate MPC directly into the decision-making loop~\cite{PETS, TDMPC, POLO, ACMPC, AC4MPC}. Related to model-based RL, early works use learned dynamics to generate model-based rollouts and improve policy learning efficiency~\cite{MBPO}; in contrast, MPC-in-the-loop approaches use learned models or value functions primarily to guide online planning rather than solely to learn a standalone policy. For example,~\cite{PETS} learns probabilistic ensemble dynamics models and leverages CEM or MPPI for online trajectory optimization.~\cite{TDMPC} jointly learns latent dynamics and value functions via temporal-difference learning and performs trajectory optimization using MPPI.~\cite{POLO} incorporates a learned value function as the terminal cost within MPC planning, while~\cite{ACMPC} constructs a cost map from RL policies to guide differentiable MPC~\cite{diff_mpc}.~\cite{AC4MPC} further initializes MPC using actor-critic estimates as terminal costs. While these approaches enhance MPC planning through learned dynamics or value functions, trajectory optimization is typically conducted using deterministic solvers or parametric sampling-based planners, rather than via non-parametric posterior refinement.

\subsection{Control as Inference with SVGD}
The control-as-inference framework formulates optimal control as probabilistic inference, where trajectory optimization corresponds to posterior inference conditioned on system dynamics and optimality variables \cite{RLCtrInf, honda2025mpcprobabilistic}. \cite{SVMPC} and \cite{DustMPC} adopt this perspective and leverage SVGD \cite{SVGD} to approximate trajectory posteriors via particle-based updates. \cite{SMPs} proposes to combine probabilistic movement primitives with SVGD to handle multi-modal trajectories for robot motion generation. \cite{Di-Plan} combines SVGD with diffusion models as informative priors and integrates path signatures to enhance diversity in the generated trajectories.\cite{S2AC} applies SVGD to approximate optimal policy distributions of RL, but focuses on single-step action updates. In contrast, Q-SVMPC extends SVGD-based inference to multi-step trajectory refinement, integrating learned policy priors and soft Q-values for non-parametric posterior updates.

\section{Preliminaries}
Our approach builds upon the SAC framework \cite{SAC} and SVMPC \cite{SVMPC}. This section briefly outlines the SAC formulation and its policy–value update scheme, followed by an introduction to MPC, its Bayesian extension, and the fundamentals of SVGD. 

\subsection{Soft Actor-Critic}
We consider an infinite-horizon Markov decision process defined by the tuple $(\mathcal{S}, \mathcal{A}, p, r, \gamma)$, where $\mathcal{S}$ and $\mathcal{A}$ denote continuous state and action spaces, $p(s_{t+1}\mid s_t,a_t)$ is the transition dynamics, $r(s_t,a_t)$ is the reward function, and $\gamma\in[0,1)$ is the discount factor.

SAC maximizes the expected cumulative reward augmented with an entropy regularization term:
\begin{equation}\label{eq:MaxEnt_Obj}
    J_\pi= 
    \mathbb{E}_{\tau\sim\pi} \left[\sum^{\infty}_{t=0}\gamma^t(r(s_t, a_t) + \alpha \mathcal{H}(\pi(\cdot|s_t))) \right],
\end{equation}
where $\alpha$ is the entropy temperature and $\tau$ denotes a trajectory induced by policy $\pi$. A higher entropy encourages more diverse actions, leading to better exploration and robustness.

Under the maximum entropy framework, the optimal policy is proportional to $\exp(\frac{1}{\alpha}Q(s,a))$. In practice, SAC updates the policy by minimizing
\begin{equation}\label{eq:sac_policy_obj}
J_\pi(\varphi) =
\mathbb{E}_{s_t\sim\mathcal{D},\,a_t\sim\pi_\varphi}
\!\left[
\alpha \log \pi_\varphi(a_t|s_t)
- Q_\theta(s_t,a_t)
\right].
\end{equation}
The soft Q-function $Q_\theta(s_t,a_t)$, parameterized by $\theta$, estimates the expected return and is optimized by minimizing the temporal-difference error:
\begingroup
\setlength{\abovedisplayskip}{3pt}
\setlength{\belowdisplayskip}{3pt}
\small
\begin{equation}\label{eq:sac_q_obj}
\begin{aligned}[t]
J_Q(\theta)
    &= \mathbb{E}_{(s_t,a_t)\sim\mathcal{D}}
       \!\left[\tfrac{1}{2}\big(Q_\theta(s_t,a_t)
       - \hat{Q}(s_t,a_t)\big)^2\right], \\[3pt]
\text{where}\quad
\hat{Q}(s_t,a_t)
    &= r(s_t,a_t)
       + \gamma\,\mathbb{E}_{s_{t+1}\sim\mathcal{D}}
       \!\left[V_{\bar{\theta}}(s_{t+1})\right], \\[3pt]
V_{\bar{\theta}}(s_{t+1})
    &= \mathbb{E}_{a_{t+1}\sim\pi}
       \!\left[Q_{\bar{\theta}}(s_{t+1},a_{t+1})
       + \alpha\,\mathcal{H}\!\left(\pi(\cdot|s_{t+1})\right)\right].
\end{aligned}
\end{equation}
\endgroup
Here, $\bar{\theta}$ denotes the parameters of the target Q-network, updated via exponential moving average.

\subsection{Bayesian Model Predictive Control}
MPC optimizes a control sequence $\mathcal{A}_t=(a_t,\dots,a_{t+H-1})$ over a finite horizon $H$ using the dynamics model $s_{t+1}=f(s_t,a_t)$. The objective is to minimize a cumulative cost over the planning horizon, and only the first action is executed in a receding-horizon manner.

Under the control-as-inference perspective \cite{RLCtrInf, SVMPC, DustMPC}, trajectory optimization can be reformulated as Bayesian inference. Let $\tau = (s_t, a_t, \dots, s_{t+H})$ denote a trajectory induced by $\mathcal{A}_t$ and dynamics $f$. We introduce an optimality variable $\mathcal{O}_\tau \in \{0,1\}$ indicating whether the trajectory is optimal. The posterior over control sequences is given by
\begin{equation}\label{eq:mpc_bayes}
p(\mathcal{A}_t \mid \mathcal{O}_\tau, s_t)
\propto
p(\mathcal{O}_\tau \mid \mathcal{A}_t, s_t)
\, p(\mathcal{A}_t \mid s_t).
\end{equation}
Here, $p(\mathcal{A}_t \mid s_t)$ represents a prior distribution over control sequences, and the likelihood is typically defined as
\begin{equation}\label{eq:liklihood_cost}
p(\mathcal{O}_\tau \mid \mathcal{A}_t, s_t)
\propto
\exp\!\big(-C(\tau)\big),
\end{equation}
where $C(\tau)$ denotes the cumulative trajectory cost. 
This formulation connects classical MPC with probabilistic inference, enabling trajectory optimization through posterior approximation.

\subsection{Stein Variational Gradient Descent}
SVGD \cite{SVGD} is a particle-based method for approximating complex posterior distributions. Given a target distribution $p(\mathcal{A})$, SVGD iteratively updates a set of particles $\{\mathcal{A}^i\}_{i=1}^M$ to minimize the KL divergence between their empirical distribution and $p$.
At each iteration, particles are updated via
\begin{equation}\label{eq:svgd_update}
    \mathcal{A}^i \leftarrow \mathcal{A}^i + \epsilon\phi^*(\mathcal{A}^i),
\end{equation}
where $\epsilon$ is the step size and the update direction is given by
\small
\begin{equation}\label{eq:svgd_phi}
    \hat{\phi}^*(\mathcal{A}^i) = 
    \frac{1}{M} \sum_{j=1}^{M}
    \left[
        \underbrace{
            k(\mathcal{A}^j,\mathcal{A}^i)
            \nabla_{\mathcal{A}^j}\log p(\mathcal{A}^j)
        }_{\text{Term \textcircled{\raisebox{-0.7pt}{1}}}}
        +
        \underbrace{
            \nabla_{\mathcal{A}^j}k(\mathcal{A}^j,\mathcal{A}^i)
        }_{\text{Term \textcircled{\raisebox{-0.7pt}{2}}}}
    \right],
\end{equation}
\normalsize
with kernel function $k(\cdot,\cdot)$. \text{Term \textcircled{\raisebox{-0.8pt}{1}}} drives particles toward high-probability regions of $p$, and \text{Term \textcircled{\raisebox{-0.8pt}{2}}} introduces repulsion to maintain particle diversity.
In Q-SVMPC, the target distribution $p(\mathcal{A})$ corresponds to the trajectory posterior induced by a learned prior and a soft Q-value–based likelihood. SVGD thus provides a non-parametric approach to approximate the trajectory posterior.

\section{Methodology}
We now present Q-SVMPC.
Sec.~\ref{sec:qstac_overview} provides an overview of the overall architecture and algorithmic flow. 
Sec.~\ref{sec:Learned Prior} introduces the RL-informed policy prior, which serves as an initialization for trajectory optimization. 
Sec.~\ref{sec:q_guidance} derives the Q-guided Stein variational inference procedure, showing how soft Q-values define an optimality likelihood within the Bayesian MPC formulation. 
Sec.~\ref{sec:entropy_comp} presents the closed-form trajectory-level entropy computation used in the learning objective. 
The complete algorithm is summarized in Alg.~\ref{algo:qstac}.

\subsection{Overview}\label{sec:qstac_overview}
Q-SVMPC formulates control as Q-guided posterior inference under a learned policy prior. As illustrated in Fig.~\ref{fig:model}, the actor learns a Gaussian prior policy and samples initial control sequence particles conditioned on the current state (Sec.~\ref{sec:Learned Prior}). These particles define a prior distribution over control sequences.

Given the current state, each particle is rolled out through the dynamics model over a finite horizon. The resulting trajectories are evaluated by the critic via soft Q-values, which define an optimality likelihood within the Bayesian MPC formulation (Sec.~\ref{sec:q_guidance}). SVGD then iteratively refines the particles to approximate the trajectory posterior, pushing them toward high-value regions while maintaining diversity.

During inference, the control sequence with the highest trajectory value is selected and its first control input is executed in a receding-horizon manner. During training, a control sequence is sampled from the refined particle set to encourage exploration, and the resulting transition is stored in the replay buffer for updating the actor and critic networks. Expectations required in the SAC objectives are approximated using empirical averages over the particle set.

\subsection{Learned Policy Distribution as Prior}
\label{sec:Learned Prior}
Within the Bayesian MPC formulation, we parameterize a prior distribution over control sequences. Specifically, we define the initial proposal distribution $q^0_t$ as a sequence of Gaussian distributions $\mathcal{N}(\mu_{t:t+H-1}, \Sigma_{t:t+H-1})$, where $H$ denotes the trajectory horizon. The mean and covariance sequences $\mu_{t:t+H-1} \triangleq (\mu_t, ..., \mu_{t+H-1})$ and $\Sigma_{t:t+H-1} \triangleq (\Sigma_t, ..., \Sigma_{t+H-1})$ are predicted by a neural network conditioned on the current state $s_t$:
$(\mu_{t:t+H-1},\Sigma_{t:t+H-1})=\pi_\varphi(s_t)$, where $\varphi$ denotes the network parameters.

A set of $M$ action-trajectory particles is then sampled from the Gaussian priors:
\begin{equation}\label{eq:sample_a0}
    \{\mathcal{A}^i\}^M_{i=1} = \{a^i_{t:t+H-1}\}^M_{i=1} \sim \mathcal{N}(\mu_{t:t+H-1}, \Sigma_{t:t+H-1}),
\end{equation}
where $\mathcal{A}$ denotes a control sequence and the superscript $i$ indexes the particle. This learned prior provides an informative initialization that is closer to the posterior distribution, thereby reducing the number of SVGD refinement steps required at execution time.

\begin{algorithm}[ht]
\caption{Q-SVMPC}
\label{algo:qstac}
\begin{algorithmic}[1]
\REQUIRE Actor $\pi_\varphi$, critic $Q_\theta$, replay buffer $\mathcal{D}$, dynamics model $f$ (or $f_\psi$), kernel $k$, horizon $H$, particles $M$
\FOR{each timestep $t$}
    \STATE Get prior $(\mu_{t:t+H-1},\Sigma_{t:t+H-1}) \leftarrow \pi_\varphi(s_t)$
    \STATE Sample particles $\{\mathcal{A}^i_t\}_{i=1}^M \sim \mathcal{N}(\mu_{t:t+H-1},\Sigma_{t:t+H-1})$
    \FOR{each SVGD step $l$ and particle $i$}
        \STATE Rollout trajectories $\tau^i = f(s_t, \mathcal{A}^i_t)$
        \STATE Compute $\hat{\phi}^*(\mathcal{A}^i_t)$ via Eq.~(\ref{eq:svgd_phi_final_term1})
        \STATE Update particles using Eq.~(\ref{eq:svgd_update})
    \ENDFOR
    \STATE Select a refined control sequence $\mathcal{A}^i_t$
    \STATE Execute first action $a_t \leftarrow \mathcal{A}^i_t[0]$ in environment
    \STATE Store transition $(s_t, a_t, r_t, s_{t+1})$ in $\mathcal{D}$
    \STATE Update $\theta$, $\varphi$ using minibatches from $\mathcal{D}$ (SAC losses with entropy in Eq.~(\ref{eq:qstac-entropy}))
\ENDFOR
\end{algorithmic}
\end{algorithm}

\subsection{Stein Variational Inference Guided by Soft Q-values}\label{sec:q_guidance}
The soft Q-value provides a natural bridge between SAC and Stein variational inference. In the following, we present a concise derivation that extends the single-step formulation in SAC to trajectory-level inference.

Given a trajectory particle $\tau^i_t = (s_t^i, a_t^i, \dots, s_{t+H}^i)$ induced by a control sequence $\mathcal{A}^i_{t:t+H-1}$, we define a trajectory score as
\begin{equation}\label{eq:Q_traj}
    S_Q(\tau^i_t) = \sum^{H-1}_{h=0}Q(s^i_{t+h},a^i_{t+h}),
\end{equation}
where $Q$ denotes the learned soft Q-function. $S_Q$ encourages predicted trajectories to remain in high-value regions throughout the rollout.

To construct a Q-guided control-as-inference likelihood, we interpret trajectory optimality as an energy-based model over rollouts. Rather than designing a hand-crafted trajectory cost $C(\tau)$, we introduce a learned energy function induced by the trajectory score:
\begin{equation}
C_S(\tau_t)\triangleq -\frac{1}{\alpha}S_Q(\tau_t).
\end{equation}
Under this definition, the optimality likelihood in Eq.~(\ref{eq:liklihood_cost}) becomes
\begin{equation}
p(\mathcal{O}_\tau \mid \mathcal{A}_t, s_t)
\propto
\exp\!\big(-C_S(\tau_t)\big)
=
\exp\!\left(\frac{1}{\alpha}S_Q(\tau_t)\right).
\end{equation}
Combining this Q-shaped likelihood with the Bayesian update in Eq.~(\ref{eq:mpc_bayes}) yields the trajectory posterior:
\begin{equation}\label{eq:optimal_post}
p(\mathcal{A}_t^i|\mathcal{O}_\tau;f,s_t) 
        \propto \exp(\frac{1}{\alpha}S_Q(\tau^i_t))q^0(\mathcal{A}_t^i;s_t),
\end{equation}
where $q^0(\mathcal{A}_t^i; s_t)$ denotes the Gaussian prior over the control sequence, as described in Sec.~\ref{sec:Learned Prior}. Accordingly, Eq.~(\ref{eq:optimal_post}) specifies the Q-shaped posterior over control sequences that Q-SVMPC seeks to approximate using SVGD.

To approximate the posterior, we adopt Stein variational inference~\cite{SVGD, SVMPC, DustMPC}. By substituting the log-posterior from Eq.~(\ref{eq:optimal_post}) into the SVGD update, we can reformulate Eq.~(\ref{eq:svgd_phi}) as
\small
\begin{align}\label{eq:svgd_phi_final_term1}
\begin{split}
    \hat{\phi}^*(\mathcal{A}_t^i, s_t) &= 
    \frac{1}{M} \sum_{j=1}^{M}\big[k(\mathcal{A}_t^j,\mathcal{A}_t^i)\nabla_{\mathcal{A}_t^j}(\frac{1}{\alpha}S_Q(\tau^j_t)+ \log q^0(\mathcal{A}_t^j;s_t)) \\ &+\nabla_{\mathcal{A}_t^j}k(\mathcal{A}_t^j,\mathcal{A}_t^i)
    \big].
\end{split}
\end{align}
\normalsize
This provides the update direction for trajectory-particle refinement, where the trajectory score provides the gradient signal that guides the SVGD particle evolution toward high-value trajectories.

\subsection{Entropy Computation}\label{sec:entropy_comp}
Q-SVMPC calculates policy entropy following S\textsuperscript{2}AC \cite{S2AC}, and further extends the formulation to the trajectory level. The resulting closed-form approximation is:
\begin{footnotesize}
\begin{align}\label{eq:qstac-entropy}
\overline{\mathcal{H}}(\pi_\varphi)
&\approx
-
\mathbb{E}_{\mathcal{A}^{0} \sim q_0}
\left[
\frac{1}{H}
\left(
\log q_0(\mathcal{A}^{0})
-
\epsilon
\sum_{l=0}^{L-1}
\operatorname{Tr}
\left(
\nabla_{\mathcal{A}^{l}}
\hat{\phi}_{l}^{*}(\mathcal{A}^{l})
\right)
\right)
\right].
\end{align}
\end{footnotesize}
Here, $\mathcal{A}^l$ denotes the particle after $l$ SVGD updates for $L$ total SVGD steps. The trace term $\operatorname{Tr}(\nabla_{\mathcal{A}^l}
\hat{\phi}_l^*(\mathcal{A}^l))$ provides a first-order approximation of the log-density change induced by the SVGD variational mapping through the change-of-variables formula.

\begin{figure}[!htbp]
    \centering
    \includegraphics[width=1.6in, height=1.3in]{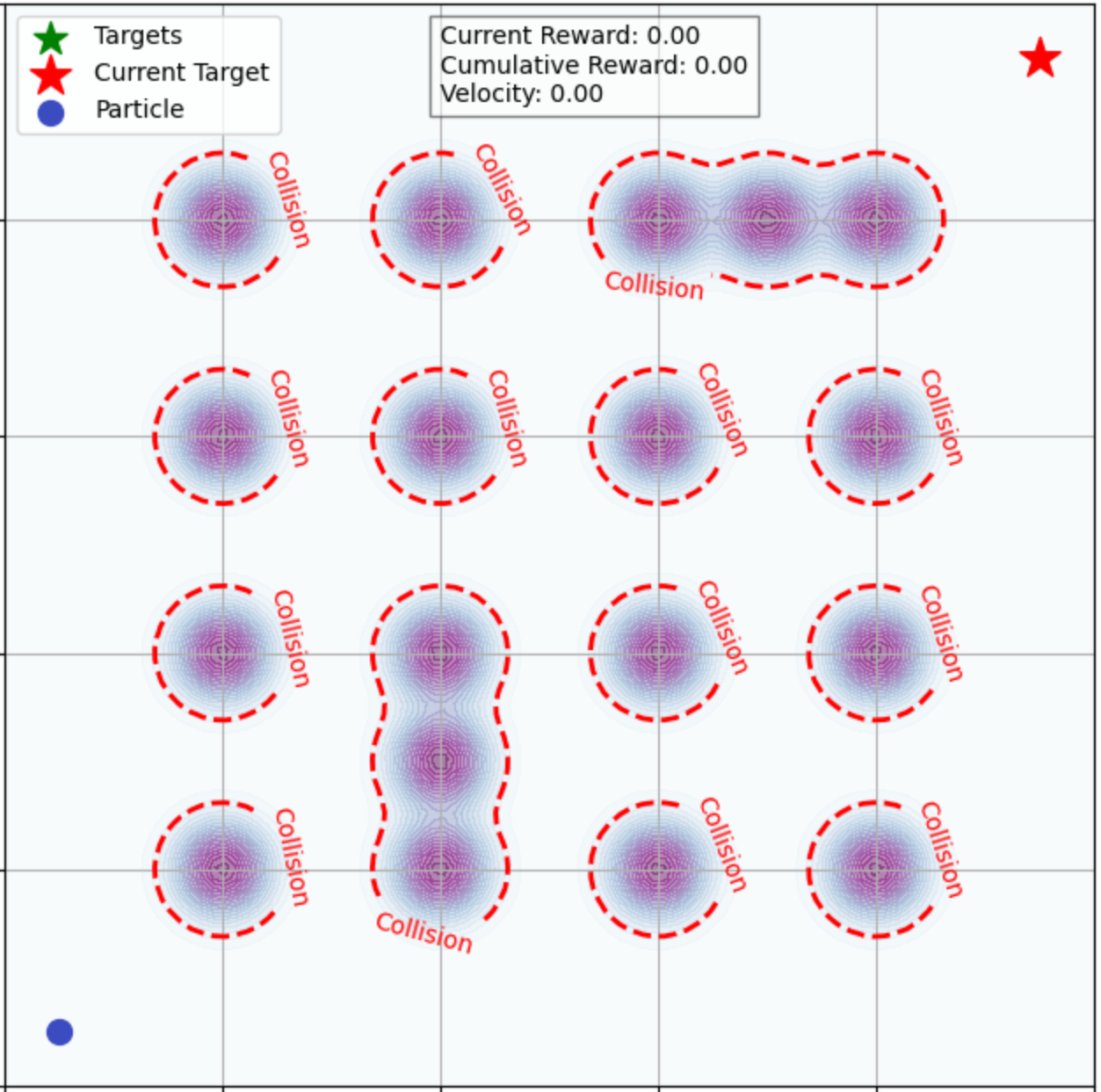}
    \includegraphics[width=1.6in, height=1.3in]{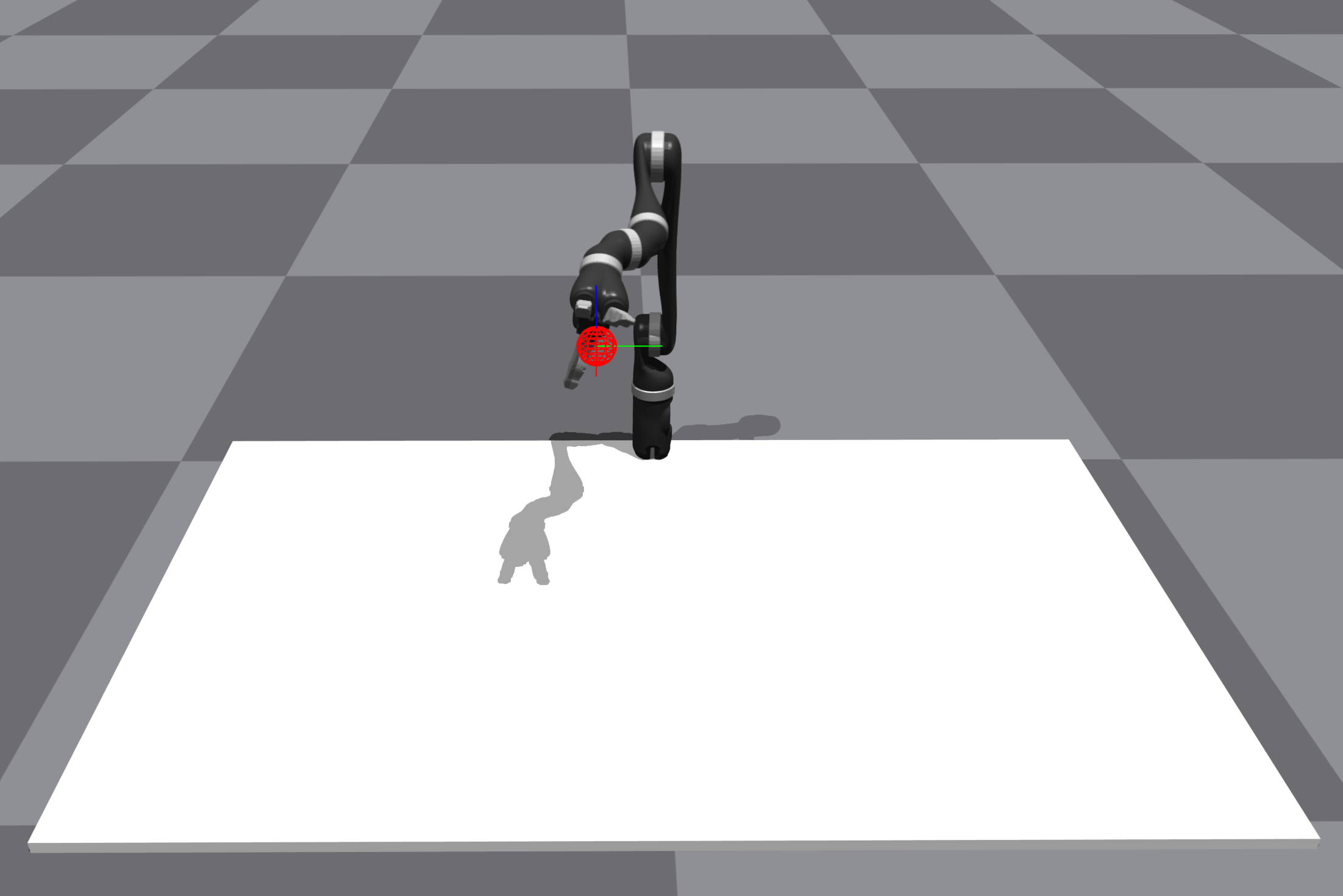}
    \includegraphics[width=1.6in, height=1.3in]{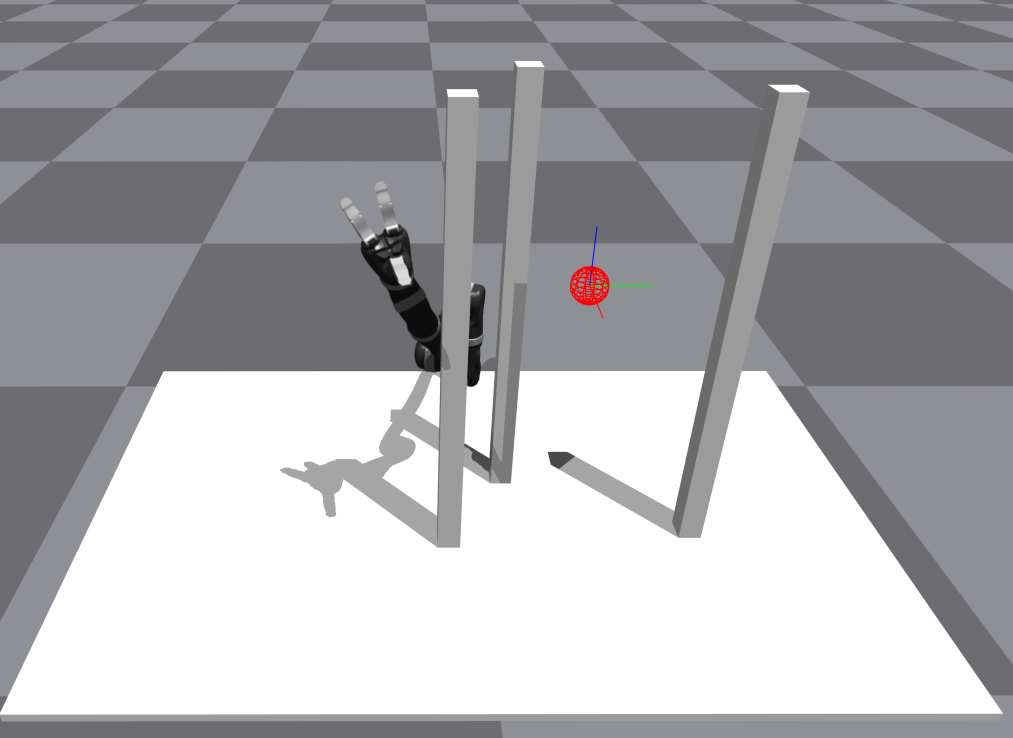}
    \includegraphics[width=1.6in, height=1.3in]{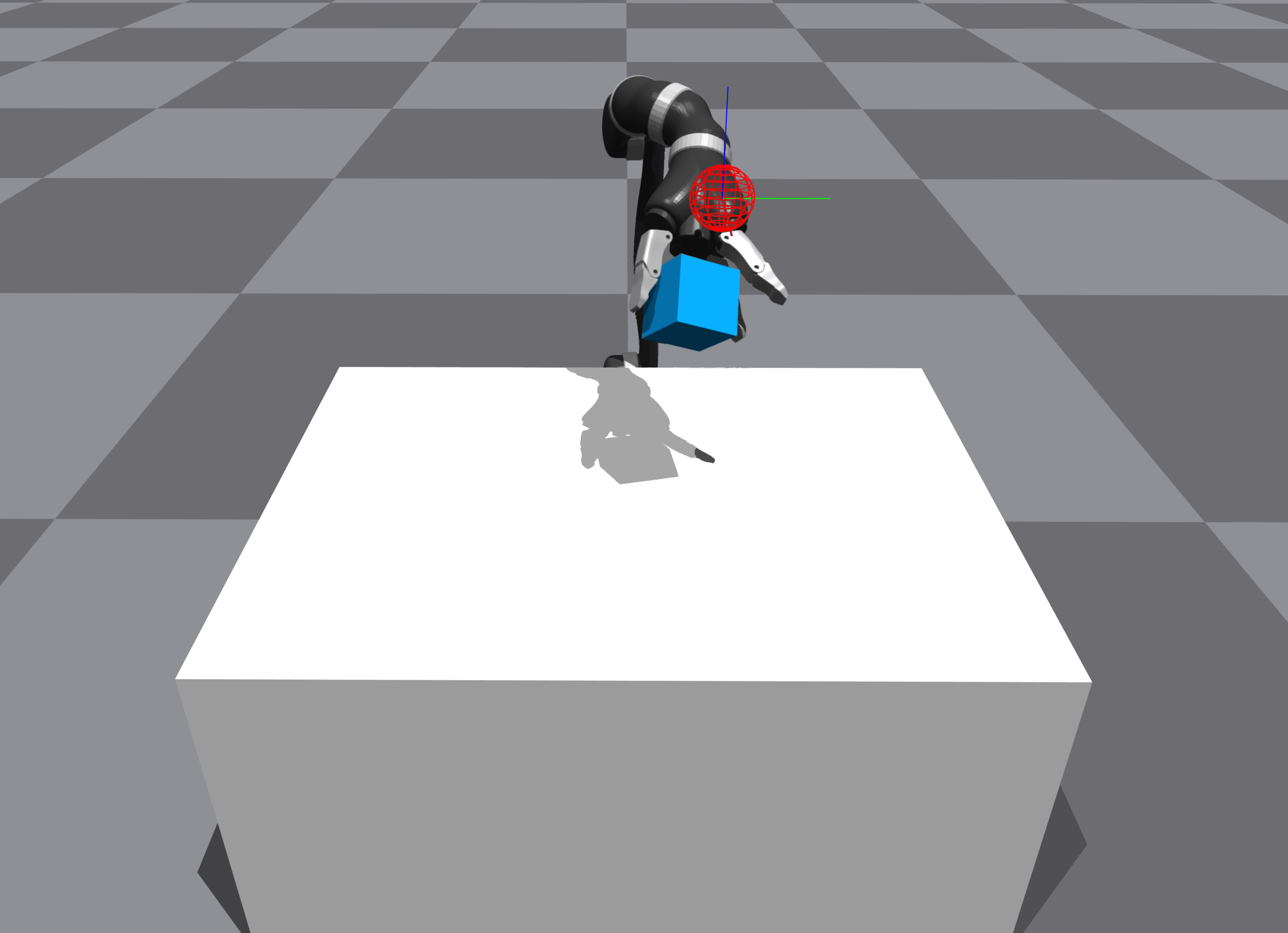}

    \caption{\textbf{Benchmark tasks used for algorithm evaluation.} 2D navigation task with multiple Gaussian obstacles (top-left); Kinova arm reaching tasks under two settings — without and with fixed obstacles (top-right/bottom-left); Kinova pick-and-place task (bottom-right).}
    \label{fig:tasks}
\end{figure}

\section{Experiments}

\subsection{Benchmarks and Setup}\label{sec:env_setup}
\subsubsection{Benchmarks}
\textbf{2D Navigation.}
We evaluate closed-loop obstacle avoidance and global planning in a 2D navigation task (Fig.~\ref{fig:tasks}, top-left), where a point mass with double-integrator dynamics must reach a fixed goal while avoiding Gaussian obstacles.

\textbf{Kinova Manipulation Suite.}
We further evaluate Q-SVMPC on robotic manipulation tasks implemented in IsaacGym~\cite{Isaacgym} using a Kinova Gen2 arm. The suite includes: \textbf{Reach}: target reaching in free space; \textbf{Reach (Obstacles)}: reaching in the presence of rigid obstacles; \textbf{Pick and Place}: grasping a cube and transporting it to a target location.

These tasks assess control performance under increasing geometric constraints and contact complexity. The robot is controlled via joint-velocity commands (6-DoF arm and 3-DoF gripper). To focus on control evaluation rather than perception, object and target states are provided directly in the state representation.

\subsubsection{Baselines}
We evaluate Q-SVMPC against \textbf{SAC}~\cite{SAC}, \textbf{S\textsuperscript{2}AC}~\cite{S2AC}, \textbf{MBPO}~\cite{MBPO}, \textbf{PETS}~\cite{PETS}, \textbf{TD-MPC}~\cite{TDMPC}, and \textbf{SVMPC}~\cite{SVMPC}, covering model-free RL, model-based RL, learning-guided MPC, and Stein variational MPC. All learning-based baselines use the same observations, action space, reward, training budget, and evaluation protocol. For TD-MPC, we use a planning-budget-matched setting with the same horizon, number of optimization iterations, and number of retained trajectories as Q-SVMPC, and implement a vectorized version for efficient training in IsaacGym~\cite{Isaacgym}. For SVMPC, we report a budget-matched variant, SVMPC\textsuperscript{$\dagger$}, and a larger-budget variant, SVMPC\textsuperscript{$\star$}.

\subsubsection{Dynamics Model}
We evaluate Q-SVMPC and planning-based baselines under both analytical $f(s, a)$ and learned dynamics models $f_\psi(s, a)$ \cite{PETS} to assess robustness to model bias. For the 2D navigation task, exact closed-form dynamics are available and used as the analytical model. For the Kinova manipulation tasks, we construct an approximate analytical model based on differentiable forward kinematics~\cite{diff_kinematics}, where future states are predicted via noisy kinematic propagation. Due to unmodeled contact effects and simplifications, this analytical model remains imperfect relative to the true system dynamics. The influence of these two model choices is evaluated through a model-type ablation study in Sec.~\ref{sec:model_type_ablation}. For SVMPC, we use the same analytical model throughout the experiments.

\begin{figure*}[!htbp]
  \centering
  \includegraphics[width=17.7cm, height=4.0cm]{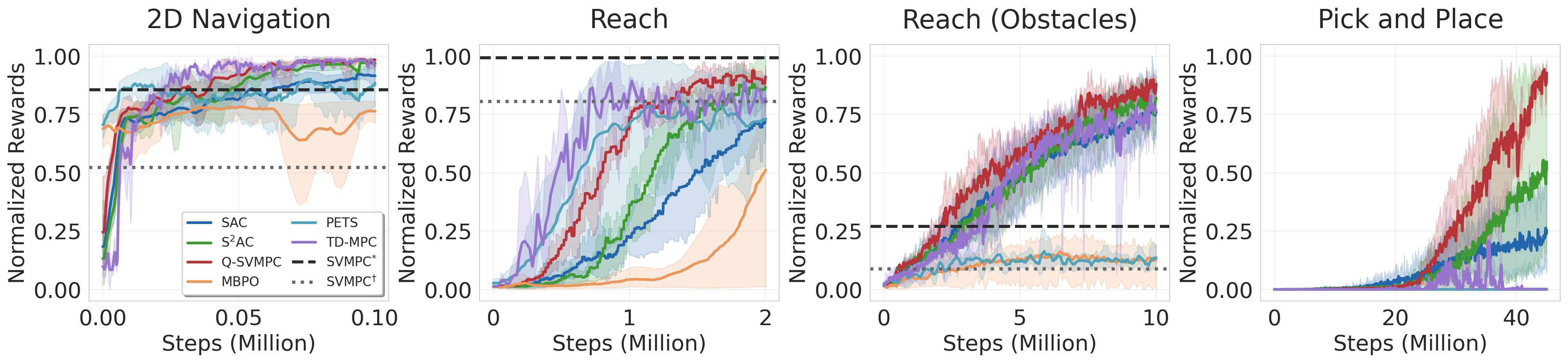}
\caption{\textbf{Training results across four benchmarks.} The figure reports the normalized cumulative return over environment interactions for SAC, S\textsuperscript{2}AC, MBPO, PETS, TD-MPC and Q-SVMPC. The x-axis denotes environment steps (in millions) and the y-axis shows normalized cumulative return. Solid curves indicate the mean over five seeds with shaded regions showing one standard deviation. Dashed and dotted horizontal lines correspond to the mean performance of two SVMPC variants (SVMPC$^{\star}$ and SVMPC$^{\dagger}$) evaluated over 20 seeds; SVMPC does not achieve successful task completion on Pick-and-Place.}
\label{fig:compare}
\end{figure*}

\begin{table}[ht!]
\caption{Success rate at different training stages (``/'' denotes not applicable or failed to converge).}
\label{tab:training_sr}
\resizebox{\columnwidth}{!}{
\begin{tabular}{@{}ccccc@{}}
\toprule
Tasks                                                   & \multicolumn{1}{l}{Algorithm} & SR@50\% & SR@75\% & SR@100\% \\ \midrule
\multicolumn{1}{c|}{\multirow{6}{*}{\begin{tabular}[c]{@{}c@{}}\\ Reach\end{tabular}}}      & SAC  & 10.5 $\pm$ 3.00 & 17.0 $\pm$ 3.50 & 72.7 $\pm$ 3.00  \\
\multicolumn{1}{c|}{}  & S\textsuperscript{2}AC         & 1.40 $\pm$ 1.10 & 13.8 $\pm$ 3.80 & 79.4 $\pm$ 5.10  \\
\multicolumn{1}{c|}{}  & MBPO                           & / & 19.7 $\pm$ 3.70 & 56.3 $\pm$ 3.90  \\
\multicolumn{1}{c|}{}  & PETS                           & 49.3 $\pm$ 18.2 & 52.8 $\pm$ 7.70 & 72.2 $\pm$ 11.6  \\
\multicolumn{1}{c|}{}  & TD-MPC                           & \textbf{76.04} $\pm$ 10.64 & \textbf{60.42} $\pm$ 44.20 & 75.00 $\pm$ 12.40  \\
\multicolumn{1}{c|}{}  & SVMPC\textsuperscript{$\dagger$}                         & / & / & \textbf{100.0} $\pm$ 0.00 \\
\multicolumn{1}{c|}{}  & SVMPC\textsuperscript{$\star$}                           & / & / & \textbf{100.0} $\pm$ 0.00 \\
\multicolumn{1}{c|}{}  & Q-SVMPC                         & 33.6 $\pm$ 3.70 & 55.0 $\pm$ 6.90 & 89.3 $\pm$ 1.90  \\ \midrule

\multicolumn{1}{c|}{\multirow{6}{*}{\begin{tabular}[c]{@{}c@{}}Reach\\ (Obstacles)\end{tabular}}} & SAC  & 9.90 $\pm$ 2.60 & 37.4 $\pm$ 4.90 & 74.7 $\pm$ 3.50  \\
\multicolumn{1}{c|}{}  & S\textsuperscript{2}AC         & 10.2 $\pm$ 2.10 & 10.9 $\pm$ 1.90 & 78.9 $\pm$ 3.50  \\
\multicolumn{1}{c|}{}  & MBPO                           & / & / & 1.70 $\pm$ 1.50  \\
\multicolumn{1}{c|}{}  & PETS                           & / & / & /  \\
\multicolumn{1}{c|}{}  & TD-MPC                           & \textbf{44.27} $\pm$ 19.09 & \textbf{54.17} $\pm$ 14.52 & 72.92 $\pm$ 10.40  \\
\multicolumn{1}{c|}{}  & SVMPC\textsuperscript{$\dagger$}                         & / & / & / \\
\multicolumn{1}{c|}{}  & SVMPC\textsuperscript{$\star$}                           & / & / & 20.0 $\pm$ 40.0 \\
\multicolumn{1}{c|}{}  & Q-SVMPC                         & 40.6 $\pm$ 3.00 & 45.4 $\pm$ 3.80 & \textbf{82.6} $\pm$ 4.40  \\ \midrule

\multicolumn{1}{c|}{\multirow{6}{*}{\begin{tabular}[c]{@{}c@{}}Pick and \\ Place\end{tabular}}}     & SAC   & / & 1.40 $\pm$ 1.20 & 80.9 $\pm$ 1.51  \\
\multicolumn{1}{c|}{}  & S\textsuperscript{2}AC         & / & 50.5 $\pm$ 7.00 & 82.7 $\pm$ 1.90  \\
\multicolumn{1}{c|}{}  & MBPO                           & / & / & /  \\
\multicolumn{1}{c|}{}  & PETS                           & / & / & /  \\
\multicolumn{1}{c|}{}  & TD-MPC                           & / & 2.34 $\pm$ 4.06 & /  \\
\multicolumn{1}{c|}{}  & SVMPC\textsuperscript{$\dagger$,$\star$}                          & / & / & / \\
\multicolumn{1}{c|}{}  & Q-SVMPC                         & 1.20 $\pm$ 0.60 & \textbf{91.4} $\pm$ 1.70 & \textbf{95.3} $\pm$ 1.10  \\ \bottomrule
\end{tabular}
}
\end{table}

\subsection{Comparative Evaluation}\label{sec:learning_results}
Our experiments evaluate whether Q-SVMPC improves learning-guided MPC in terms of sample efficiency, training stability, and robustness across benchmarks. We further analyze collision avoidance via constraint-violation metrics (Sec.~\ref{sec:constraint}) and qualitatively examine the diversity of the refined trajectories.

Fig.~\ref{fig:compare} compares the training performance of Q-SVMPC and the baselines across four tasks. On 2D Navigation, Q-SVMPC attains the highest mean final return, while TD-MPC and S\textsuperscript{2}AC achieve comparable performance. On \textit{Reach}, TD-MPC learns rapidly during the early stage of training, and SVMPC$^{\star}$ achieves the highest final return, while Q-SVMPC reaches the best final performance among the learning-based policy methods. On \textit{Reach (Obstacles)}, Q-SVMPC obtains the highest mean final return. On \textit{Pick-and-Place}, Q-SVMPC converges reliably, whereas TD-MPC and the other learned model-based baselines do not converge within the training budget, and SVMPC does not achieve successful task completion under its evaluation budget. Overall, TD-MPC and PETS show strong early-stage learning on the simpler tasks. However, TD-MPC produces less consistent results across seeds and shows larger fluctuations during training, while Q-SVMPC provides more consistent performance across the four benchmarks.

As a more interpretable metric for manipulation, Tab.~\ref{tab:training_sr} reports task success rates on the Kinova suite at 50\%, 75\%, and 100\% of training progress. For learning-based methods, we report the mean and standard deviation across five independent training seeds. Each checkpoint is evaluated using 10 episodes with 100 parallel environments. We count a success if the end-effector stays within 10\,cm of the target for 20 steps (reaching) or reaches within 5\,cm after grasping for 30 steps (pick-and-place). Since SVMPC is an inference-time planner, we only report SR@100\% averaged over 20 seeds and mark SR@50\%/SR@75\% as ``/''.

Tab.~\ref{tab:training_sr} shows that Q-SVMPC achieves strong and reliable success across tasks, with particularly large gains on the more challenging settings. SVMPC attains 100\% success on \textit{Reach}, reflecting effective trajectory planning in free space; however, its performance degrades as task complexity increases, even with a longer horizon and more particles. In contrast, Q-SVMPC leverages an RL-informed policy prior and Q-value guidance, incorporating experience rather than planning from scratch, which leads to substantially higher success rates on obstacle-rich reaching and contact-intensive manipulation. TD-MPC is more sample-efficient during the early stages of training on \textit{Reach} and \textit{Reach (Obstacles)}, but also exhibits greater inconsistency across seeds, particularly on \textit{Reach}. On \textit{Pick-and-Place}, Q-SVMPC achieves a high and consistent success rate, while the evaluated planning-based and model-based baselines fail to converge under the available computational and training budgets. Overall, these results indicate that combining an RL-informed policy prior with Q-guided non-parametric SVGD refinement provides consistent performance across tasks of increasing geometric and contact complexity.

\subsection{Empirical Collision–Return Trade-off}
\label{sec:constraint}
In Q-SVMPC, we leverage the trajectory score to reduce the collision rate.
Tab.~\ref{tab:safety_performance} summarizes the collision--performance
trade-off on 2D Navigation and \textit{Reach (Obstacles)}, reporting the
collision rate (the fraction of collision timesteps over all timesteps;
lower is better) together with the episodic return (higher is better).
On 2D Navigation, Q-SVMPC achieves the highest episodic return
($-206.60$), while TD-MPC records a lower collision rate
(1.17\% versus 5.49\%) with a slightly lower return ($-219.36$).
S\textsuperscript{2}AC tends to take paths with a higher collision
frequency in exchange for higher returns, while
SVMPC\textsuperscript{$\star$} avoids all observed collisions but obtains
a lower return ($-684.5$), illustrating the trade-off between conservative
obstacle avoidance and task performance. On \textit{Reach (Obstacles)},
where rigid-body obstacles block the direct path, Q-SVMPC achieves both
the lowest observed collision rate (2.42\%) and the highest return (74.9),
compared with 5.23\% and 61.3 for TD-MPC.
SVMPC\textsuperscript{$\dagger$} maintains a low collision rate (2.87\%)
but achieves a low return (7.06), suggesting conservative behavior without
informative guidance. With a larger planning budget,
SVMPC\textsuperscript{$\star$} more frequently enters the blocked region,
resulting in both a higher collision rate and a higher return. In contrast,
Q-SVMPC achieves a higher return without an increase in the observed
collision rate.
  
\begin{figure}[!ht]
    \centering
    \includegraphics[width=0.49\linewidth]{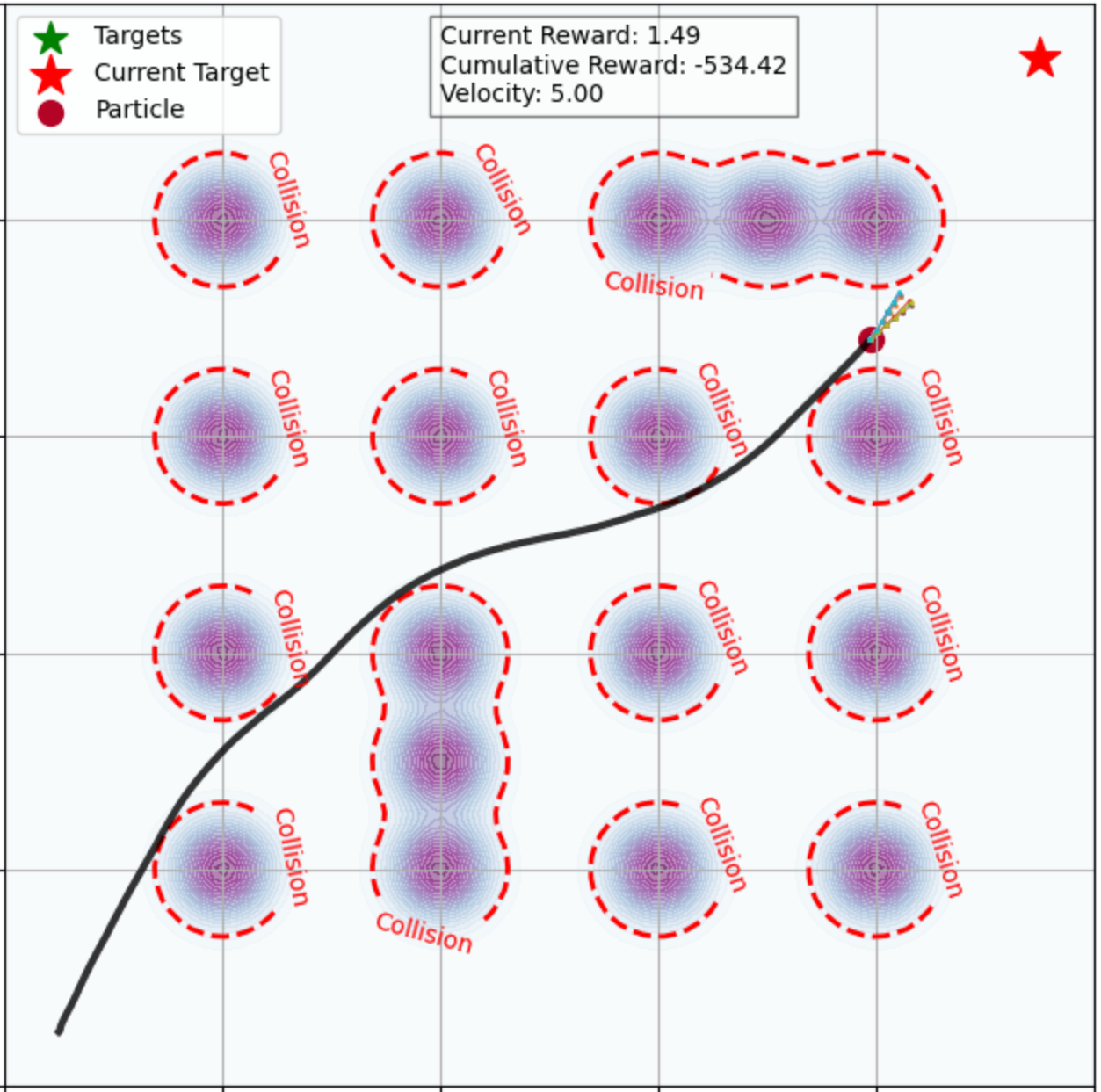}
    \includegraphics[width=0.49\linewidth]{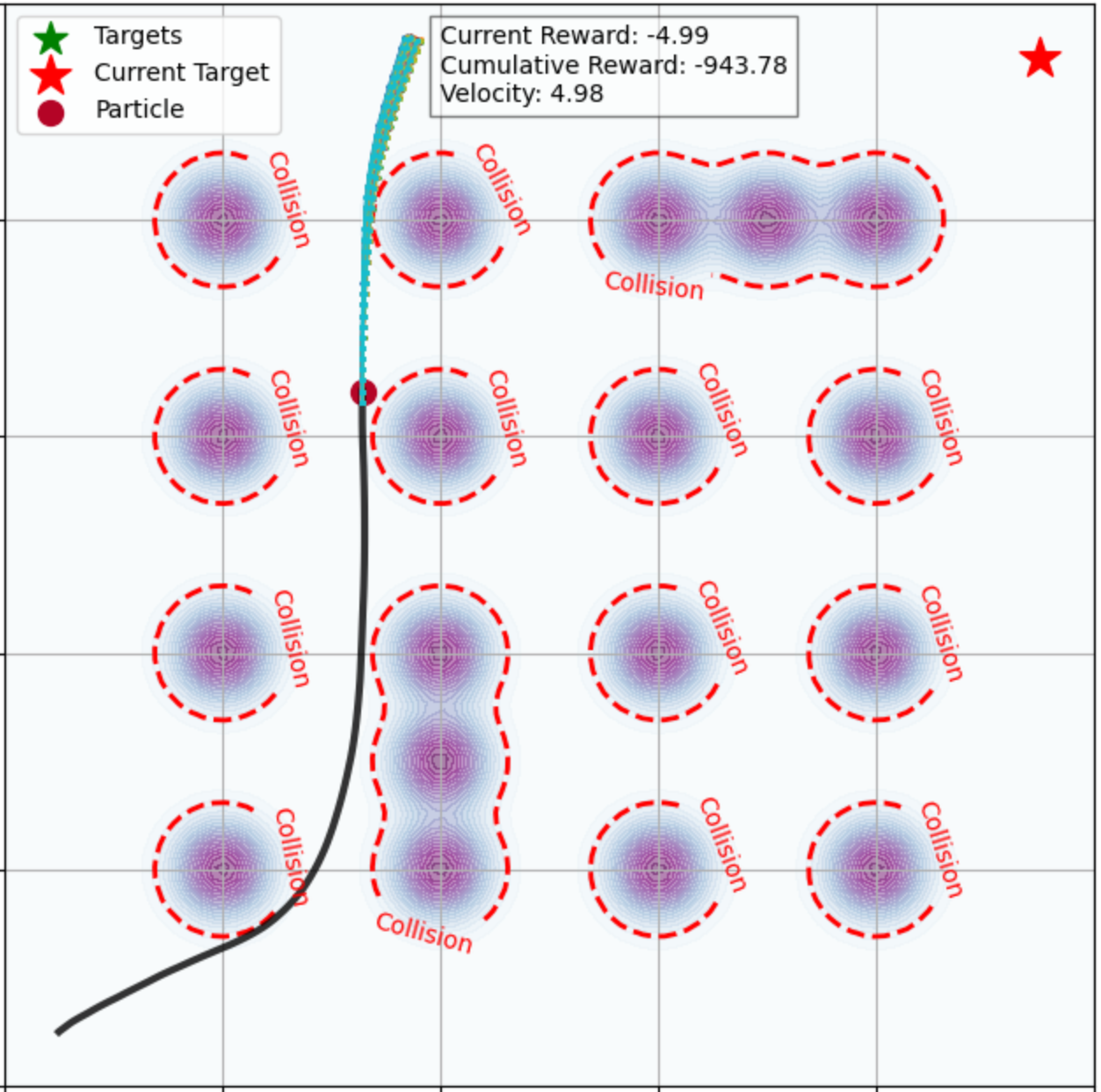}
    \caption{\textbf{Trajectory visualization comparison.} Executed and sampled trajectories on 2D Navigation for Q-SVMPC (left) and SVMPC\textsuperscript{$\star$} (right).}
    \label{fig:traj_vis}
\end{figure}

Fig.~\ref{fig:traj_vis} qualitatively illustrates this trade-off. SVMPC\textsuperscript{$\star$} follows a more conservative route around the obstacles, whereas Q-SVMPC selects a narrower, higher-return route closer to the obstacle boundaries. This behavior is consistent with the use of an informative policy prior and Q-value guidance during trajectory refinement.

\begin{table}[!ht]
\centering
\caption{Collision Rate and Episodic Return.}
\label{tab:safety_performance}
\resizebox{\columnwidth}{!}{
\begin{tabular}{lcc|cc}
\toprule
\multirow{2}{*}{\textbf{Method}} 
& \multicolumn{2}{c|}{\textbf{2D Navigation}} 
& \multicolumn{2}{c}{\textbf{Reach (Obstacles)}} \\
& Coll. (\%) $\downarrow$ & Return $\uparrow$ 
& Coll. (\%) $\downarrow$ & Return $\uparrow$ \\
\midrule
SAC    & 28.2 & -719.12 & 8.35 & 60.8 \\
S\textsuperscript{2}AC   & 10.1 & -371.30 & 4.03 & 63.4 \\
MBPO   & 11.0 & -632.18 & 27.9 & 14.6 \\
PETS   & 4.50 & -1265.33 & 27.6 & 7.77 \\
TD-MPC & 1.17 & -219.36 & 5.23 & 61.3 \\
SVMPC\textsuperscript{$\dagger$}  & 22.5 & -1998.4 & 2.87 & 7.06 \\
SVMPC\textsuperscript{$\star$}  & \textbf{0.00} & -684.5 & 20.9 & 21.4 \\
Q-SVMPC & 5.49 & \textbf{-206.60} & \textbf{2.42} & \textbf{74.9} \\
\bottomrule
\end{tabular}
}
\end{table}

Overall, TD-MPC achieves faster early-stage learning on the reaching tasks and a lower collision rate on 2D Navigation, whereas Q-SVMPC obtains the highest return on 2D Navigation and the strongest observed collision--return combination on \textit{Reach (Obstacles)}. These results suggest that Q-guided non-parametric trajectory refinement is beneficial in obstacle-rich and contact-intensive settings, while introducing a task-dependent trade-off between performance, collision frequency, and computational cost.

\subsection{Ablation Study}\label{sec:abla}
To further assess Q-SVMPC's design choices, we perform ablations on the policy prior, planning horizon, and dynamics-model type.

\begin{figure}[!ht]
    \centering
    \includegraphics[width=0.49\linewidth]{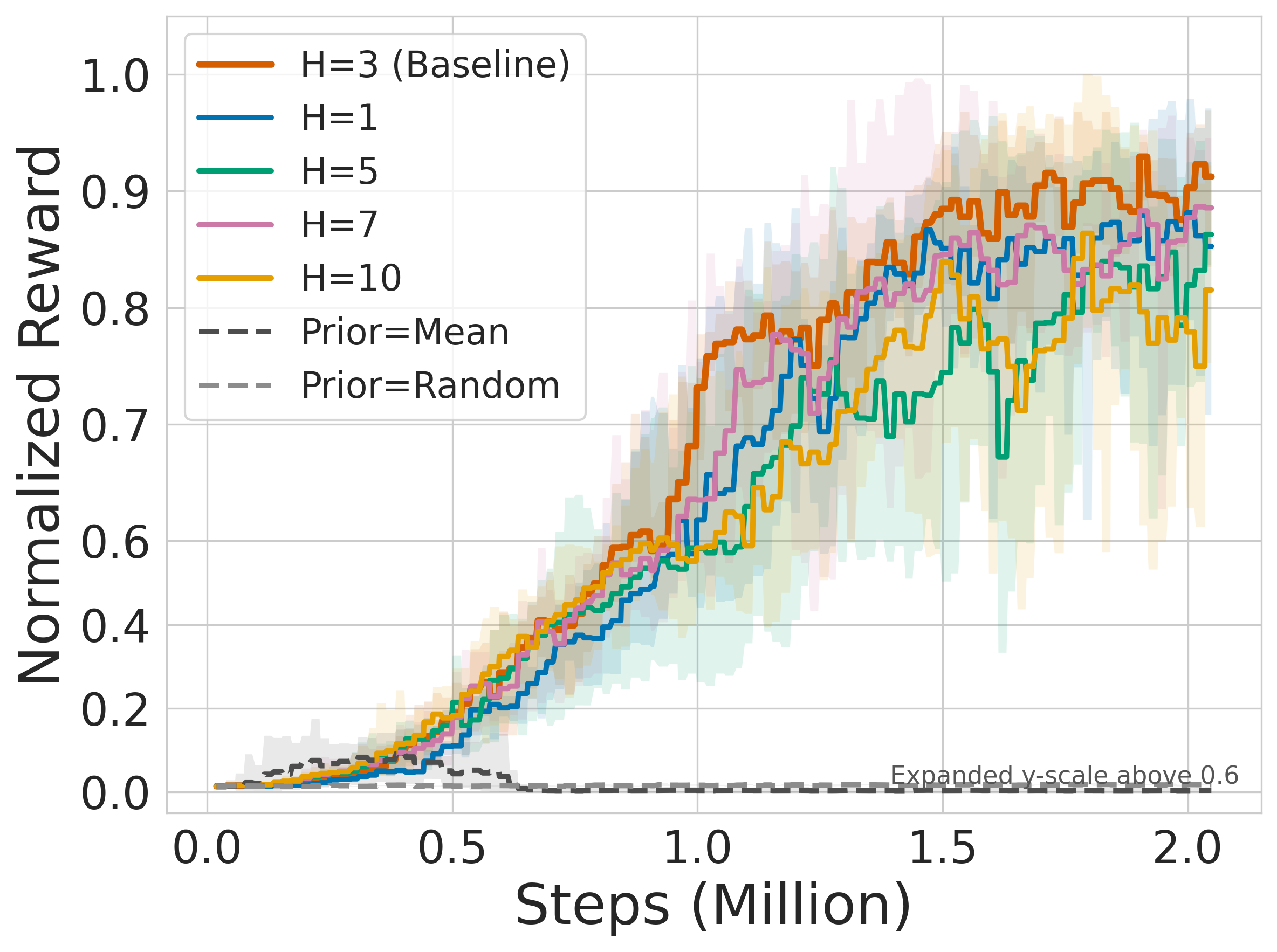}
    \includegraphics[width=0.49\linewidth]{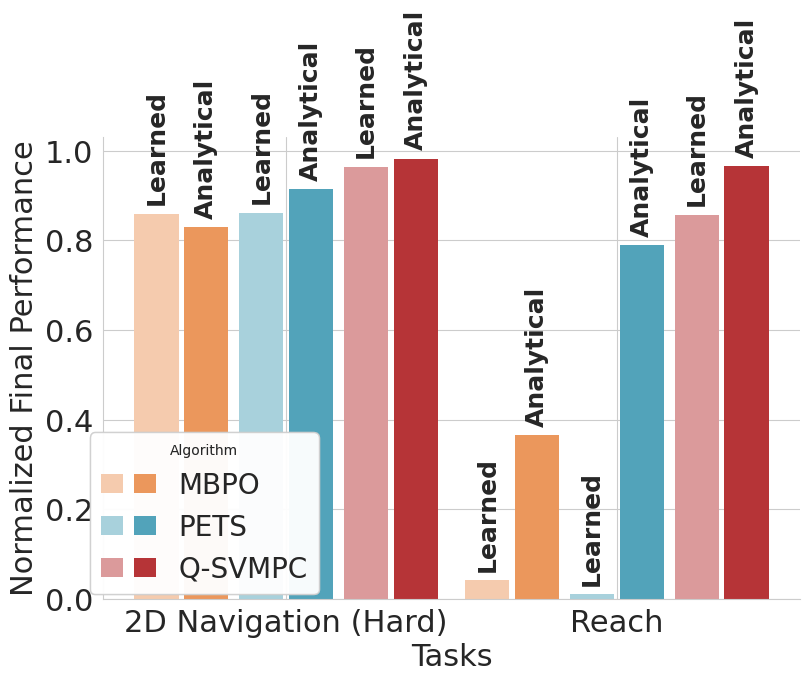}
    \caption{\textbf{Ablation results.} Prior selection and horizon length on the \textit{Reach} task with Q-SVMPC (left); analytical vs. learned dynamics on \textit{2D Navigation} and \textit{Reach (Obstacles)} for model-based methods (right). }
    \label{fig:ablation_two}
\end{figure}

\subsubsection{Prior Type}
Fig.~\ref{fig:ablation_two}(left) shows the training curves of Q-SVMPC in the \textit{Reach} task under different prior distributions. The dashed lines correspond to two alternative priors: Random, which samples initial particles from a Gaussian distribution at every action step, and Mean, which initializes the particles using the mean action predicted by the current policy. For fairness, both variants use a fixed horizon of 3, and we increase their SVGD update steps and particle count to strengthen their optimization capability.

As illustrated in the figure, the Random prior fails to converge, while the Mean prior provides an early improvement but degrades during later training. These results demonstrate that the SAC-learned prior provides a more effective initialization under the same planning budget.

\subsubsection{Horizon Length}
The solid curves in Fig.~\ref{fig:ablation_two}(left) compare different planning horizons on the \textit{Reach} task. $H=3$ achieves the best overall performance and maintains a higher mean return than $H=1$, indicating that short multi-step lookahead provides useful guidance beyond one-step Q-value optimization. The gain is moderate because \textit{Reach} is a smooth, obstacle-free task with a direct path to the target. Longer horizons do not provide further improvement, likely due to accumulated model error and increased optimization difficulty.

\subsubsection{Analytical and Learned Dynamics Models}\label{sec:model_type_ablation}
We evaluate Q-SVMPC under both an analytical dynamics model $f(s,a)$ and a learned model $f_\psi(s,a)$~\cite{PETS} to assess robustness to model bias. For 2D navigation, we use the exact closed-form dynamics; for Kinova, we use an approximate analytical model based on differentiable forward kinematics with noisy propagation~\cite{diff_kinematics}, which is imperfect due to unmodeled contact effects. The model-type ablation (Sec.~\ref{sec:model_type_ablation}) shows that Q-SVMPC retains similar performance under analytical and learned dynamics on 2D Navigation. On \textit{Reach (Obstacles)}, the learned model leads to a small reduction in final performance relative to the analytical model, although Q-SVMPC remains competitive with the evaluated model-based baselines. These results indicate that Q-SVMPC remains sensitive to dynamics-model accuracy, but exhibits greater tolerance to model error than the other evaluated model-based baselines.

\subsection{Computational Cost}\label{sec:compute}
Tab.~\ref{tab:inference_time} reports the per-step inference time, control frequency, and planning parameters measured on an RTX 4060 laptop, with all methods evaluated on \textit{2D Navigation}. For PETS and TD-MPC, the particle count is reported as n/N, where N candidate trajectories are sampled and the top n elites are retained at each planning iteration. The iteration column denotes SVGD update steps for S\textsuperscript{2}AC, SVMPC, and Q-SVMPC, and planning iterations for PETS and TD-MPC. TD-MPC, SVMPC\textsuperscript{$\dagger$}, and Q-SVMPC use the same horizon of 5 and three optimization iterations, with 10 particles. Q-SVMPC runs at 24.6ms (40.7Hz), compared with 14.7ms for TD-MPC and 15.5ms for SVMPC\textsuperscript{$\dagger$}. Its additional latency arises from backpropagating through the dynamics rollouts during SVGD refinement. PETS is the slowest under its recommended sampling configuration, while the larger-budget SVMPC\textsuperscript{$\star$} requires 79.9ms (12.5Hz). Overall, Q-SVMPC remains practical for online control despite the additional gradient-computation cost.

\begin{table}[t]
\centering
\caption{Inference time per control step and equivalent control frequency with planning-related parameters.}
\label{tab:inference_time}
\resizebox{\columnwidth}{!}{
\begin{tabular}{lccccc}
\toprule
\textbf{Method} 
& \textbf{Horizon} 
& \textbf{Particles} 
& \textbf{Iterations} 
& \textbf{Time (ms) $\downarrow$}
& \textbf{Frequency (Hz) $\uparrow$} \\
\midrule
SAC     & -  & -   & - & 0.296 & 3378.4 \\
S\textsuperscript{2}AC    & -  & 10  & 3 & 13.6  & 73.5   \\
PETS    & 15 & 35/350 & 5 & 578   & 1.73   \\
TD-MPC  & 5 & 10/100 & 3 & 14.7   & 67.9   \\
SVMPC\textsuperscript{$\dagger$} & 5  & 10  & 3 & 15.5  & 64.3    \\
SVMPC\textsuperscript{$\star$}   & 50  & 100  & 3 & 79.9  & 12.5    \\
Q-SVMPC & 5  & 10  & 3 & 24.6  & 40.7   \\
\bottomrule
\end{tabular}
}
\end{table}

\subsection{Sim-to-Real}
\begin{figure}[!ht]
    \centering
    \includegraphics[width=0.49\linewidth]{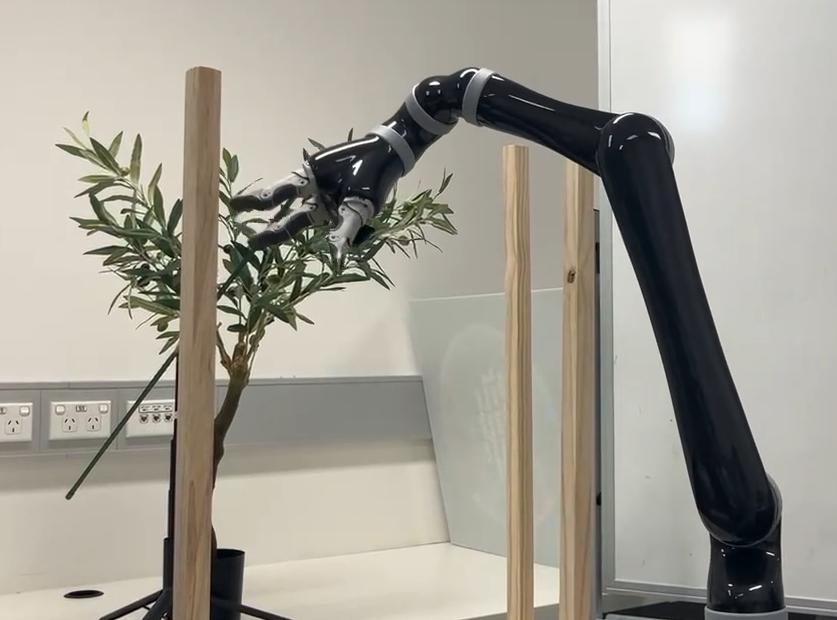}
    \includegraphics[width=0.49\linewidth]{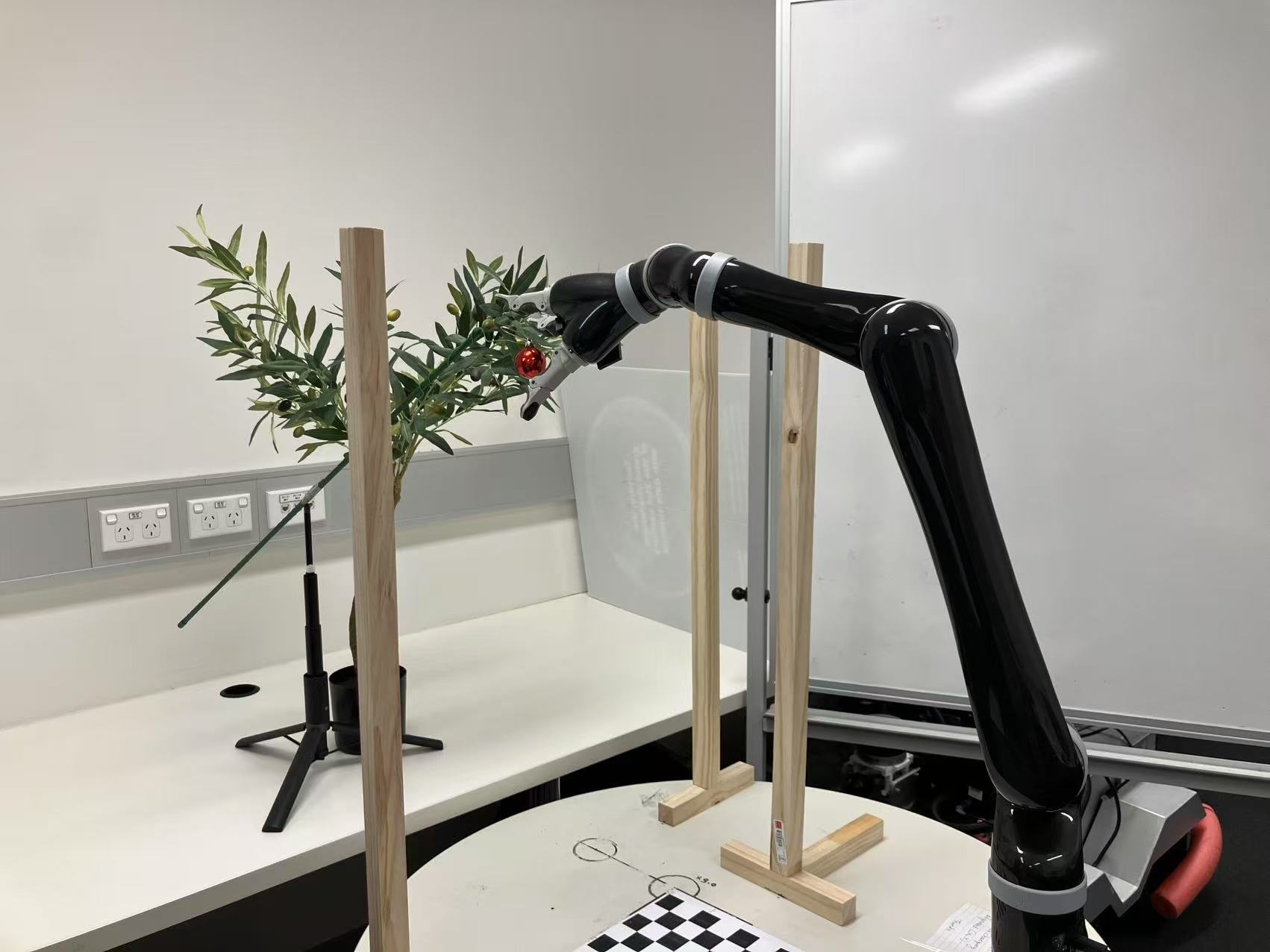}
    \caption{\textbf{Real-world deployment on the Kinova arm.}  The robot performs obstacle avoidance (left) and reaches the target fruit under closed-loop control by Q-SVMPC.}
    \label{fig:real_scene}
\end{figure}

\subsubsection{General Setup}
We train models on a desktop machine equipped with an NVIDIA RTX 4090 GPU as well as on laptops with RTX 4060 GPUs. A separate workstation runs the Kinova ROS API, which we interact with through a custom REST service over Ethernet. The Kinova arm is controlled via joint velocities, with a planner rate of 40 Hz and inner-loop streaming at 60 Hz using $n$ sub-steps.

\subsubsection{Adaptive Action Integration for Robot Control}
To mitigate the reality gap between simulation and real-world execution, we adopt a modified action integration strategy inspired by~\cite{PCAP}. Mismatches between simulation and hardware (e.g., friction, damping, and stiffness) can accumulate steady-state errors during policy transfer. We therefore integrate high-level action commands over time to improve robustness.

Following~\cite{PCAP}, we employ an adaptive action integrator to interface the high-level planner with the Kinova ROS controller running at a higher servo rate. At each control step $t$, the planner outputs a target joint-velocity command $a_t$ (i.e., the first action of the optimized trajectory conditioned on $s_t$). To avoid abrupt changes and to match the inner-loop frequency, we up-sample this command into $n$ intermediate updates. Specifically, given the previously executed command $u_t$, we compute the per-segment increment:
\begin{equation}
\delta a_t^{d}=\frac{a_t-u_t}{n},
\end{equation}
where $n$ is the segmentation factor. The desired command is then integrated over the $n$ sub-steps as
\begin{equation}
u_{t,k+1}=u_{t,k} + \delta a_t^{d},\quad k=0,\ldots,n-1,\quad u_{t,0}=u_t,
\end{equation}
and we send $u_{t,k}$ sequentially to the Kinova controller. This integration yields smooth transitions between successive MPC updates while maintaining a high-rate command stream for stable real-world execution.

\subsubsection{Real Experiments}
We evaluate Q-SVMPC on a real Kinova arm performing obstacle-aware fruit picking using a policy directly transferred from the \textit{Reach (Obstacles)} task. At each step, Q-SVMPC generates joint-velocity commands online; once the end-effector reaches the target, the gripper is closed manually. Fig.~\ref{fig:real_scene} shows the setup and representative rollouts, while Tab.~\ref{tab:real_sr} reports results over 15 trials for SAC, S\textsuperscript{2}AC, and Q-SVMPC. Other simulation baselines, including TD-MPC, were not evaluated on hardware.

Q-SVMPC achieves the highest success rates in both obstacle avoidance and target reaching. SAC frequently collides with obstacles, while S\textsuperscript{2}AC avoids them more reliably but often misses the target. These results demonstrate robust obstacle-aware control under real-world effects such as unmodeled friction, joint backlash, and sensor latency.

\begin{table}[]
\caption{Obstacle-aware reaching for fruit picking}
\label{tab:real_sr}
\begin{tabular}{@{}cccc@{}}
\toprule
Task                                                                                & Algorithm & \multicolumn{2}{c}{Success Rate (\%)} \\ \midrule
\multirow{6}{*}{\begin{tabular}[c]{@{}c@{}}Real World\\ Picking Fruit\end{tabular}} & \multicolumn{1}{l}{} & \multicolumn{1}{l}{Avoiding Obstacles} & \multicolumn{1}{l}{Reaching Target} \\ 
& SAC   & 20.0  & 40.0      \\
& S\textsuperscript{2}AC & 86.7 & 60.0 \\
& Q-SVMPC & \textbf{93.3} & \textbf{80.0}  \\ \bottomrule
\end{tabular}
\end{table}

\section{Conclusion and Future Work}
We presented Q-SVMPC, a learning-guided MPC approach that casts trajectory optimization as Bayesian inference with an RL-informed policy prior and soft Q-values as the optimality signal, and approximates the resulting posterior using SVGD. This inference view connects MPC with soft value learning, enabling non-parametric trajectory refinement that preserves particle diversity and provides a principled alternative to hand-crafted cost shaping. Across the evaluated benchmarks, Q-SVMPC achieves the highest final performance on \textit{Reach (Obstacles)} and \textit{Pick-and-Place}, while remaining competitive with SVMPC and TD-MPC on the simpler \textit{Reach} and 2D Navigation tasks. Ablations further validate the roles of the learned prior, horizon length, and robustness to model inaccuracies. For future work, we will extend Q-SVMPC to vision-based settings by incorporating visual observations into both value learning and dynamics modeling, enabling planning under partial observability and more complex scene geometry.


\bibliographystyle{IEEEtran}
\bibliography{references}

\end{document}